\documentclass[conference]{IEEEtran}
\ifCLASSOPTIONcompsoc
  \usepackage[nocompress]{cite}
\else
  \usepackage{cite}
\fi
\ifCLASSINFOpdf
\else
\fi
\newcommand{\citep}[1]{\cite{#1}}
\usepackage{url}
\usepackage{booktabs}
\usepackage{amsfonts}
\usepackage{amsmath}
\usepackage{amssymb}
\usepackage{graphicx}

\newcommand{\cescore}[1]{\mathrm{CE}^{\mathrm{score}}_{#1}}

\newcommand{\cepos}[1]{\mathrm{CE}^{\mathrm{pos}}_{#1}}

\begin{document}
%
% paper title
% Titles are generally capitalized except for words such as a, an, and, as,
% at, but, by, for, in, nor, of, on, or, the, to and up, which are usually
% not capitalized unless they are the first or last word of the title.
% Linebreaks \\ can be used within to get better formatting as desired.
% Do not put math or special symbols in the title.
\title{Plausible Counterfactual Explanations \\of Recommendations}

% author names and affiliations
% use a multiple column layout for up to three different
% affiliations
\author{\IEEEauthorblockN{Jakub Černý}
\IEEEauthorblockA{Dateio\\
Prague, Czechia\\
Email: jakub.cerny@dateio.eu}
\and
\IEEEauthorblockN{Jiří Němeček}
\IEEEauthorblockA{Faculty of Electrical Engineering\\
Czech Technical University in Prague\\
Prague, Czechia\\
Email: contact@nemecekjiri.cz}
\\
\IEEEauthorblockN{Jakub Mareček}
\IEEEauthorblockA{Faculty of Electrical Engineering\\
Czech Technical University in Prague\\
Prague, Czechia\\
Email: jakub.marecek@fel.cvut.cz}
\and
\IEEEauthorblockN{Ivan Dovica}
\IEEEauthorblockA{Dateio\\
Prague, Czechia\\
Email: ivan.dovica@dateio.eu}
}

% conference papers do not typically use \thanks and this command
% is locked out in conference mode. If really needed, such as for
% the acknowledgment of grants, issue a \IEEEoverridecommandlockouts
% after \documentclass

% for over three affiliations, or if they all won't fit within the width
% of the page (and note that there is less available width in this regard for
% compsoc conferences compared to traditional conferences), use this
% alternative format:
%
%\author{\IEEEauthorblockN{Michael Shell\IEEEauthorrefmark{1},
%Homer Simpson\IEEEauthorrefmark{2},
%James Kirk\IEEEauthorrefmark{3},
%Montgomery Scott\IEEEauthorrefmark{3} and
%Eldon Tyrell\IEEEauthorrefmark{4}}
%\IEEEauthorblockA{\IEEEauthorrefmark{1}School of Electrical and Computer Engineering\\
%Georgia Institute of Technology,
%Atlanta, Georgia 30332--0250\\ Email: see http://www.michaelshell.org/contact.html}
%\IEEEauthorblockA{\IEEEauthorrefmark{2}Twentieth Century Fox, Springfield, USA\\
%Email: homer@thesimpsons.com}
%\IEEEauthorblockA{\IEEEauthorrefmark{3}Starfleet Academy, San Francisco, California 96678-2391\\
%Telephone: (800) 555--1212, Fax: (888) 555--1212}
%\IEEEauthorblockA{\IEEEauthorrefmark{4}Tyrell Inc., 123 Replicant Street, Los Angeles, California 90210--4321}}

% use for special paper notices
%\IEEEspecialpapernotice{(Invited Paper)}

% make the title area
\maketitle

% As a general rule, do not put math, special symbols or citations
% in the abstract
\begin{abstract}
Explanations play a variety of roles in various recommender systems, from a legally mandated afterthought, through an integral element of user experience, to a key to persuasiveness.
A natural and useful form of an explanation is the Counterfactual Explanation (CE). We present a method for generating highly plausible CEs in recommender systems and evaluate it both numerically and with a user study.
\end{abstract}

% no keywords

% For peer review papers, you can put extra information on the cover
% page as needed:
% \ifCLASSOPTIONpeerreview
% \begin{center} \bfseries EDICS Category: 3-BBND \end{center}
% \fi
%
% For peerreview papers, this IEEEtran command inserts a page break and
% creates the second title. It will be ignored for other modes.
\IEEEpeerreviewmaketitle

\section{Introduction}
Explanations in recommender systems (RS) have been repeatedly shown to improve user experience \citep{herlocker_explaining_2000,shang_why_2022} by enhancing trust in the RS, user satisfaction with the RS, and transparency of the system \citep{nunes_systematic_2017}. Recommender explanations (or ``recsplanations'' \citep{mcinerney_explore_2018}) also measurably increase the persuasiveness of the RS \citep{nunes_systematic_2017}
and help users make decisions faster.
Shang et al. \citep{shang_why_2022} showed that around half of users (up to 62\% in e-commerce) are \textit{highly interested} in obtaining an explanation.
In addition, regulation \citep{gdpr,CCPA,ai-act} requires explanations for automated decisions made using personal data.

There are many forms of explanations \citep{ge_survey_2024} of RS. Here, we focus on Counterfactual Explanations (CEs,  \citep{wachter_counterfactual_2017}) in particular.
% what are they?
Counterfactual explanations, in general, answer the question ``How would the input have to change for the output to change a given way?'' For example: ``In order to recommend Kill Bill, your rating of Pulp Fiction would have to be at least 4 stars out of 5.''
Counterfactual explanations are well understandable and considered useful by users \citep{shang_why_2022} and even suggested\footnote{A direct citation: ``[...] it is sufficiently transparent and intelligible to inform the data subject of the extent to which a variation in the personal data taken into account would have led to a different result.''} by recent EU court judgment
\citep{gdpr_supports_ce_case}. This makes them well-suited for practical use.
% \jirimargincomment{neco z toho why not paperu, consider taky From intrinsic to counterfactual: On the explainability of contextualized recommender systems}
Counterfactual explanations are an active area of research outside of recommender systems \citep{guidotti_counterfactual_2022} but also within the RS domain \citep{ge_survey_2024, baklanov_ceers_2024}.

% However, one crucial aspect is somewhat omitted in the literature on CEs for recommenders. This aspect
One may wish for the CE to be actionable (e.g., not recommending becoming younger or less educated) and sparse (i.e., utilizing as few changes as possible) and these qualities remain the same across a variety of application domains. However, there is an overlooked desideratum of \textit{plausibility}, which is rarely addressed by CE methods for recommenders. In agreement with the commonly considered definition \citep{guidotti_counterfactual_2022}, we define plausibility as the likelihood of the counterfactual with respect to the training data.
Let us consider the same Tarantino example as above, with the current user having rated Pulp Fiction 4 stars and asking about the recommendation of Kill Bill. A counterfactual ``Kill Bill would not be recommended, if you rated Pulp Fiction 3 stars.'' is more plausible than ``Kill Bill would not be recommended, if you rated Pulp Fiction 1 star'', because it is more likely, conditional on the attribute values of the present user of the RS.
While both counterfactuals are perfectly valid, %but assuming that few similar users with such an out-of-distribution rating exist, this is unlikely to be applicable. A
a user then might consider the latter explanation less useful, since such a low rating is outside of the margin for error in the ratings, due to mood at the time of rating or watching the movie or similar.
%\jirimargincomment{an actual example would be better here}

There are several general-purpose methods for the generation of CE with plausibility objectives. We build on LiCE \citep{nemecek2025generating}, which has recently been shown to outperform other comparable methods.
Our contributions are as follows:
\begin{itemize}
    \item We propose a modification of LiCE as a method for plausible Counterfactual Explanations of recommendations.
    % \item We specify of
    \item We compare various setups on the plausibility of CEs.
    \item We validate the proposed modification with a user study.
\end{itemize}
With our contributions, we open the question of plausible explanations of Recommender Systems, evaluate the top-performing model, and describe the pros and cons of each setup. Our work enables practitioners to make informed decisions in order to improve both the user experience and the performance of their system \citep{nunes_systematic_2017}.

% ------------- outline ---------------
% - explanations are useful
% - in multiple dimensions (persuasiveness etc.)
% - well-studied in general  postponed to related works
% - counterfactuals are great (define them)
% - supported by regulation (link to the court decision)
% - studied little in this domain? NOPE
% - quite popular, (recent surveys)
% - special focus on plausibility
% - similar arguments as in lice [FIND an example where the plausible CE is clearly better otherwise make one up]
% - there are many methods, but the recent best, with support for binary vars and tweakable to one's liking, is LiCE
% - we evaluate it [on more data???? and more methods????], considering a user study

% \jiricomment{our contributions - listed clearly - explain why relevant for RecSys people...}

\section{Preliminaries}
% \paragraph{Notation}
% We consider the following problem.
Let there be $D$ items. Let $x_i \in \mathcal{X}$ be the vector of interactions (e.g., clicks, views, ratings...) of a user $i$, where $\mathcal{X} \subseteq \mathbb{R}^D$ is the $D$-dimensional space of such interactions. Each user-item interaction is a scalar that can take real values (e.g., length of an interaction), but they are oftentimes integer or binary (e.g., item ratings or purchases). We have a set $\mathcal{D} = \{x_1, x_2, \ldots , x_N\}$ of $N$ such vectors, representing $N$ users. We refer to this set as the (training) dataset.

\paragraph{LiCE}
% with support for binary vars and tweakable to one's liking, is LiCE
We modify LiCE \citep{nemecek2025generating}, a method for generating Likely Counterfactual Explanations (CEs). LiCE utilizes on Mixed-Integer Optimization (MIO, \cite{wolsey_integer_2021}), a popular framework for solving NP-Hard optimization problems  \cite{kochProgressMathematicalProgramming2022}, and tractable probabilistic models \cite{poon_sum-product_2011,NEURIPS2023_6b61c278,sidheekh2023probabilistic}, a popular framework for modelling uncertainty, somewhat resembling neural networks in the compositionality of the model. In particular, LiCE utilizes a Sum-Product Network (SPN,  \cite{poon_sum-product_2011}) to estimate the likelihood of a counterfactual. SPNs and MIO are particularly suitable for the use in RS, where the input variables are often discrete, which MIO and SPNs model naturally.

\paragraph{SPN}
To elaborate: SPNs are a class of tractable probabilistic models, taking the form of a directed acyclic graph (DAG), where each node represents a distribution over a defined set of variables (item interactions, in our case). The leaf nodes in the DAG represent simple distributions, often univariate. The non-leaf nodes are of one of two types. There are sum nodes, which produce a weighted mixture of the distributions in the ``successor nodes''. The remaining non-leaf nodes are product nodes, which produce the product of distributions in the ``successor nodes''.

% However, they mostly ommit Counterfactual Explanations (CEs), with the exception of the work by Shang et al. \citep{shang_why_2022}, which does not propose a method for generating CEs, but rather validates the usefulness and user preference of counterfactuals.

\section{Related works}
A quarter century of work on explaining recommendations since the pioneering work by Herlocker et al. \citep{herlocker_explaining_2000} addressed the question of explaining black-box recommenders. The 25 years of development and the diverse landscape of explaining recommender systems is described by a multitude of reviews \citep{mcsherry_explanation_2005,nunes_systematic_2017}, taxonomies \citep{ friedrich_taxonomy_2011,nunes_systematic_2017}, surveys \citep{vultureanu-albisi_survey_2022,wardatzky_evaluating_2024,ge_survey_2024}.
%even PhD theses \citep{melchiorre_explainable_2024}.
We shall focus on the Counterfactual Explanations, best surveyed recently by Ge et al. \citep{ge_survey_2024}, and well motivated by Shang et al. \citep{shang_why_2022}, based on user studies and interviews.

\paragraph{Generating Counterfactual Explanations in Recommender Systems.} Let us mention several key methods: PRINCE \citep{ghazimatin_prince_2020}, focused on finding CEs over a heterogeneous information network, assuming that personalized PageRank is used to infer recommendations.
ACCENT \citep{tran_counterfactual_2021} provides CEs for recommenders based on neural networks, but it only considers a limited number of items and seeks a change of the top-ranked one, rather than considering the entire ranking.
% taky maj nic moc fidelity?
%
CR-VAE \citep{xu_learning_2021} utilizes a Variational Auto-Encoder (VAE) to obtain CEs and then finds causal dependencies.
% co je s timhle spatne, to nevim :D ~97% validity (asi), taky plausible... maybe shit proximity??? - jestli rozumim ~20 edits? - nope, in latent space - would have to do my own?
%
PGCE \citep{he_prototype-guided_2023} also utilizes VAE. It uses gradient-based methods, with an additional loss term on the distance to a prototype (mean of almost recommended items) in the latent space. The authors evaluate ``realism'', but they measure it as normalized $\ell_2$-distance from the factual, while we propose utilizing a probabilistic model.
CountER \citep{tan_counterfactual_2021} focuses on the items being recommended (rather than user-item interactions), proposing similar counterfactual items that would no longer be recommended.
CF$^2$ \citep{Tan_2022} and GREASE \citep{chen_grease_2022} are methods of generating CEs for recommenders based on GNNs.
Finally, Kaffes et al. \citep{kaffes_model-agnostic_2021} generate model-agnostic CEs by search, and Zhong and Negre \citep{zhong_shap-enhanced_2022} expand on that by ordering the interactions used in the search according to their SHAP values.
We refer to \cite{ge2024survey} for a recent survey.

\paragraph{The contrast.}
Our approach improves on the state of the art by reliably finding the closest counterfactual explanations for most recommender models without the need of exhaustive search. The generated CEs have a high likelihood with respect to the data distribution and are faithful to the model.
% \jirimargincomment{are others less plausible? we'll never know}
We obtain these by modifying LiCE \citep{nemecek2025generating}, a well-performing method for generating plausible counterfactuals and applying it to recommender systems. Our modification is also openly accessible at \url{https://github.com/Epanemu/LiCE4Recommenders}.
We know of no other publicly implemented method that constrains CEs by removing an item from top-$k$ recommendations, which is what we find desirable.

% \jirimargincomment{briefly discuss: MIO - in recommender systems?, SPNs for recmmender systems?

% show that plausibility is omitted...
% }

\section{Our Method}
% \jirimargincomment{check if the other models next to ease are also representable using MIO - discuss that here through "best guess" claims.}
To model the counterfactual generation for ranking as a Mixed-Integer Optimization (MIO) problem, we have to formulate the ranking process to ensure validity. Throughout, we assume the availability of a scoring function $f_j(x_i)$ for each item $j$ and select the top-$k$ items with the highest values to be recommended.
This holds for Neural Collaborative Filtering (NCF, \citep{he_neural_2017}), EASE \cite{steck_embarrassingly_2019}, and even more recent VASP \citep{vancura_deep_2021}, for example.
% Given that EASE is just a matrix multiplication, formulating the prediction score $y_i$ of a merchant $i$ is
Subsequently, the MIO formulation reads
\begin{align}
    \min & \sum_{j=1}^{D} |x_{i,j} - x_{i,j}^{\prime}| \label{eq:objective} \\
    % \text{such that} \; & y_j = B_{j,:} \cdot x^{\prime} & \forall j \in \{1, \ldots, n\} \\
    \text{such that} \; & y_j = f_j(x_i^{\prime}) & \forall j \in \{1, \ldots, D\} \\
    & x^{\prime}_{i,j} \in \{0, 1\}, y_j \in \mathbb{R} & \forall j \in \{1, \ldots, D\},
\end{align}
where $x_i$ is the factual (the input vector of user $i$) and $x_i^{\prime}$ is the counterfactual. The absolute value can be readily linearized. In the binary case, $x_j - x_{i,j}^{\prime}$ if $x_{i,j} = 1$ and $x_{i,j}^{\prime} - x_{i,j}$ if $x_{i,j} = 0$. In the continuous case, one can introduce \cite{wolsey_integer_2021,nemecek2025generating} auxiliary variable. Note that we slightly diverge from LiCE in using a plain $\ell_1$ distance. This is because we mainly work with binary variables.

We must also be able to formulate the scoring function $f_j$ within MIO. For Neural models \citep{he_neural_2017} including VAEs \citep{liang_variational_2018}, we refer to the widely studied topic of representing Neural Networks in MIO (e.g., \citep{anderson_strong_2020}), where ready-made implementations exist \citep{bynum_pyomo_2021}.
For EASE \citep{steck_embarrassingly_2019} (and similarly for SLIM \citep{ning_slim_2011}), $f_j(x^{\prime})$ is simply $B_{j,:} \cdot x^{\prime}$, where $B_{j,:}$ is the $j$-th row of the matrix $B$.
Even more recent high-performing models, such as H+ Vamp Gated \citep{kim_enhancing_2019} or VASP
% a combination of multiple Neural architectures
\citep{vancura_deep_2021} could be formulated, using approximations or non-linearity in the formulation.

To avoid the trivial optimum $x_i^{\prime} := x_i$, we constrain the value $y_c$ of the item $c$, for which we wish to find the explanation.
Naively, we could define a counterfactual $\cescore{\tau}$ with the minimal change required for the score of merchant $i$ to drop below a threshold $\tau$:
\begin{equation}
    y_i \le \tau.
\end{equation}
Unfortunately, this does not ensure that item $c$ is not be recommended, given the counterfactual. It might happen that the counterfactual change would decrease the scores of all items, and the ordering would remain the same. To establish what makes item $c$ rank high, we define a counterfactual $\cepos{\rho}$, which is the minimal change in $x_i$, such that the item $c$ is among the top-$\rho$ items.
In the MIO formulation, we introduce $D-1$ binary variables $r_j$, which equal 1 if and only if the merchant $j$ ranks above the merchant $i$. Formulating the constraints
\begin{align}
    y_c & \le y_j + (1 - r_j) * M & \forall j \in \{1,\ldots,D\} \setminus \{c\} \\
    \sum_{j = 1}^D r_j & \ge \rho, \label{eq:rank}
\end{align}
where $M$ is a ``big-M'' constant \citep{wolsey_integer_2021} large enough to cover the span of possible $y_c$ values. It can be computed for each $j$ by subtracting the minimal value of $y_j$ from the maximal possible value of $y_c$. We want the big-M to be as low as possible, since a high $M$ can cause numerical issues. The constraint \eqref{eq:rank} ensures that $\rho$ items have a score greater than or equal to that of the item $c$. This corresponds to a rank of at least $\rho + 1$.

% Finally, we might want to know how we could achieve some specific ordering of a subset of $m$ items.
% % For example, we wish to know why an item $i$ is ranked higher than comparable items $j$ and $k$. We thus ask what would need to happen so that the merchant $j$ was first, then $k$ and the merchant $i$ was last.
% % \jirimargincomment{make up a proper example usecase or just remove this}
% We define a counterfactual $\ceorder{O}$, requiring that merchants follow the ordering $O = (o_1, \ldots o_m)$, meaning that $y_{o_l} \ge y_{o_{l+1}}$ for all $l \in \{1, \ldots, m-1\}$.
% % In the previous example, $O = (j, k, i)$.
% And this is precisely how we formulate the constraints.
% We formulate this as
% \begin{equation}
%     y_{o_l} \ge y_{o_{l+1}} \quad \forall l \in \{1, \ldots, m-1\}.
% \end{equation}

\subsection{SPN variants}
By default, LiCE utilizes SPNs modeled on all features, i.e., considers each item separately. While having the best granularity, one might need a lot of data to capture the joint distribution without overfitting. It might also lead to SPNs, which are costly to optimize over. Luckily, items often belong to some categories or come with tags. Let there be $K$ categories, each described by a set $\mathcal{K}_j$, which contains items belonging to the category $j$. We can then aggregate the sparse information into $K$-dimensional feature vectors and use those for training.

Denote the resulting vector of aggregated values for user $i$ by $z_i$. We evaluate three aggregating functions. The first aggregating function is the sum, which can be formulated in MIO directly by
\begin{equation}
  z_{i,j} = \sum_{k \in \mathcal{K}_j} x_{i, k}^{\prime} \quad \forall j \in \{1, 2, \ldots, K\}.
\end{equation}
The second aggregating function is a mean, which is modeled the same way, but multiplied by the inverse of the number of items in a given category
\begin{equation}
  z_{i,j} = \frac{1}{|\mathcal{K}_j|} \sum_{k \in \mathcal{K}_j} x_{i, k}^{\prime} \quad \forall j \in \{1, 2, \ldots, K\}.
\end{equation}
Utilizing the mean is fundamentally different from the sum in that the sum retains the discrete property of input data, while the mean leads to non-discrete random variables, which are handled differently in the SPN.
Finally, there is the disjunction (logical or) modeled by
\begin{align}
  z_{i,j} \le & \sum_{k \in \mathcal{K}_j} x_{i, k}^{\prime} & \forall j \in \{1, 2, \ldots , K\} \label{eq:force_0} \\
  z_{i,j} \ge & \frac{1}{|\mathcal{K}_j|} \sum_{k \in \mathcal{K}_j} x_{i, k}^{\prime} & \forall j \in \{1, 2, \ldots , K\} \label{eq:force_1} \\
  z_{i,j} \in & \{0, 1\} & \forall j \in \{1, 2, \ldots , K\},
\end{align}
where we assume all $x_{i,k}^{\prime}$ are binary, the constraint \eqref{eq:force_0} forces the binary $z_{i,j}$ to be zero when all items in a given category are not interacted with (have value 0). In all other cases, the constraint \eqref{eq:force_1} will force $z_{i,j}$ to equal one. This formulation can be extended to non-binary inputs by introducing appropriate scaling factors.

\section{Numerical Evaluation}

% \subsection{Setup}

\begin{table*}[t]
    \centering
    \caption{Plausible CE within a time limit: proportion of
    % globally optimal
    results obtained within the time limit.
    Only for 8.6\% of CEs, their global optimality could not be proven within the time limit of
   10 minutes. Median run-time was 11 seconds.}
    \resizebox{\textwidth}{!}{\begin{tabular}{llcccccccccc}
    \toprule
    % \multicolumn{2}{l}{Variants}
    & & \multicolumn{4}{c}{Optimize ($\alpha = 0.1$)} & \multicolumn{4}{c}{Threshold (median LL)} & \multicolumn{2}{c}{No SPN} \\
    \cmidrule(lr){3-6} \cmidrule(lr){7-10} \cmidrule{11-12}
    % \multicolumn{2}{l}{Input formats}
    & & \multicolumn{3}{c}{Binarized input} & Rating  & \multicolumn{3}{c}{Binarized input} & Rating & Binarized input & Rating \\
    \cmidrule(lr){3-5} \cmidrule(lr){6-6}  \cmidrule(lr){7-9} \cmidrule(lr){10-10} \cmidrule{11-11} \cmidrule(lr){12-12}
    % \multicolumn{2}{l}{Aggregator functions}
    & Validity variant  & Disjunction & Sum & Mean & Mean & Disjunction & Sum & Mean & Mean & N/A & N/A \\
    \cmidrule(lr){1-2}
    \cmidrule(lr){3-12}
    % \cmidrule(lr){3-6} \cmidrule(lr){7-10} \cmidrule(lr){11-12}
    Amazon & $k$-th rank & 1.00 & 0.99 & 0.97 & 0.97 & 0.92 & 1.00 & 0.00 & 0.00 & 1.00 & 1.00 \\
    & score threshold & 1.00 & 0.95 & 1.00 & 1.00 & 0.94 & 0.95 & 0.00 & 0.00 & 1.00 & 1.00 \\

Yelp & $k$-th rank & 1.00 & 1.00 & 0.30 & 0.07 & 0.86 & 0.00 & 0.00 & 0.00 & 1.00 & 1.00 \\
& score threshold & 1.00 & 1.00 & 0.95 & 0.88 & 1.00 & 0.00 & 0.00 & 0.00 & 1.00 & 1.00 \\

Netflix & $k$-th rank & 1.00 & 1.00 & 0.70 & 0.51 & 0.99 & 0.00 & 0.00 & 0.33 & 1.00 & 1.00 \\
& score threshold & 1.00 & 1.00 & 0.89 & 0.62 & 1.00 & 0.00 & 0.00 & 0.33 & 1.00 & 1.00 \\
% \midrule
% Dateio - categories & $k$-th rank  & 1.00 & 1.00 & 1.00 & - & 0.58 & 1.00 & 1.00 & - & 1.00 & - \\
%  & score threshold  & 1.00 & 1.00 & 1.00 & - & 0.58 & 1.00 & 1.00 & - & 1.00 & - \\
% Dateio - tags & $k$-th rank  & 1.00 & 1.00 & 1.00 & - & 0.30 & 1.00 & 0.00 & - & 1.00 & - \\
%  & score threshold  & 1.00 & 1.00 & 1.00 & - & 0.12 & 1.00 & 0.00 & - & 1.00 & - \\
    \bottomrule
    \end{tabular}}
    \label{tab:rates}
\end{table*}

\begin{table*}[t]
    \centering
    \caption{A tradeoff between plausibility and similarity
    evaluated as log-likelihood and $\ell_1$ distance: mean and standard deviation over all retrieved CEs of a given configuration. %Following results from Table \ref{tab:rates}, we do not show results of sum and mean aggregator functions for binarized input results and rating results for thresholding variant, due to low success rates. These results suggest a tradeoff between plausibility and similarity.
    }
    \resizebox{\textwidth}{!}{\begin{tabular}{llcccccccccc}
    \toprule
    & & \multicolumn{5}{c}{\textbf{Plausibility (Log-Likelihood)}} & \multicolumn{5}{c}{\textbf{Similarity ($\ell_1$ Distance)}} \\
    \cmidrule(lr){3-7} \cmidrule(lr){8-12}
    & & \multicolumn{3}{c}{Binarized input - Disjunction} & \multicolumn{2}{c}{Rating - Mean} & \multicolumn{3}{c}{Binarized input - Disjunction} & \multicolumn{2}{c}{Rating - Mean} \\
    \cmidrule(lr){3-5} \cmidrule(lr){6-7}  \cmidrule(lr){8-10} \cmidrule(lr){11-12}
    & & Optimize & Threshold & No SPN & Optimize & No SPN & Optimize & Threshold & No SPN & Optimize & No SPN \\
    \cmidrule(lr){1-7} \cmidrule(lr){8-12}
    Amazon & $k$-th & $-18.44 \pm 12.23$ & $-12.08 \pm 3.16$ & $-19.57 \pm 13.45$ & $1695 \pm 31$ & $1693 \pm 34$ & $1.16 \pm 0.49$ & $2.60 \pm 3.24$ & $1.08 \pm 0.43$ & $2.11 \pm 2.16$ & $0.66 \pm 0.38$ \\
    & score & $-18.40 \pm 11.93$ & $-12.21 \pm 3.42$ & $-19.61 \pm 13.42$ & $1695 \pm 35$ & $1694 \pm 34$ & $1.19 \pm 0.46$ & $2.82 \pm 3.94$ & $1.11 \pm 0.37$ & $2.33 \pm 3.19$ & $0.54 \pm 0.37$ \\
    Yelp & $k$-th & $-89.11 \pm 38.68$ & $-78.33 \pm 13.72$ & $-103.21 \pm 46.04$ & $3185 \pm 113$ & $3122 \pm 148$ & $2.71 \pm 2.04$ & $4.16 \pm 7.69$ & $1.65 \pm 0.98$ & $2.54 \pm 1.98$ & $1.07 \pm 0.86$ \\
    & score & $-89.23 \pm 39.80$ & $-82.42 \pm 14.69$ & $-103.06 \pm 45.99$ & $3184 \pm 84$ & $3122 \pm 148$ & $2.44 \pm 1.57$ & $5.02 \pm 10.89$ & $1.62 \pm 0.89$ & $6.92 \pm 5.17$ & $1.01 \pm 0.79$ \\
    Netflix & $k$-th & $-23.44 \pm 9.10$ & $-19.67 \pm 2.89$ & $-23.61 \pm 9.25$ & $381 \pm 44$ & $367 \pm 60$ & $0.71 \pm 1.03$ & $4.07 \pm 8.34$ & $0.69 \pm 0.90$ & $0.53 \pm 1.13$ & $0.39 \pm 0.73$ \\
    & score & $-23.33 \pm 9.11$ & $-19.66 \pm 2.86$ & $-23.62 \pm 9.18$ & $385 \pm 45$ & $368 \pm 60$ & $2.07 \pm 1.37$ & $4.87 \pm 5.57$ & $2.07 \pm 1.37$ & $1.84 \pm 4.01$ & $1.28 \pm 1.12$ \\
 %    \cmidrule(lr){1-7} \cmidrule(lr){8-12}
 %     Dateio - categs & nth & $-7.91 \pm 3.03$ & $-5.91 \pm 0.64$ & $-7.93 \pm 3.02$ & - & - & $1.00 \pm 0.00$ & $1.00 \pm 0.00$ & $1.00 \pm 0.00$ & - & - \\
 % & score & $-7.91 \pm 3.03$ & $-5.91 \pm 0.64$ & $-7.92 \pm 3.03$ & - & - & $1.00 \pm 0.00$ & $1.00 \pm 0.00$ & $1.00 \pm 0.00$ & - & - \\
 % Dateio - tags & nth & $-310.5 \pm 56.6$ & $-83.40 \pm 0.39$ & $-372.8 \pm 70.2$ & - & - & $4.96 \pm 1.87$ & $87.02 \pm 18.69$ & $1.00 \pm 0.00$ & - & - \\
 % & score & $-302.7 \pm 59.5$ & $-83.64 \pm 0.04$ & $-366.9 \pm 61.9$ & - & - & $4.75 \pm 2.17$ & $92.17 \pm 15.33$ & $1.00 \pm 0.00$ & - & - \\
    \bottomrule
    \end{tabular}}
    \label{tab:stats}
\end{table*}

We evaluate our method on three public datasets: Yelp \citep{yelp_data}, Amazon book reviews \citep{amazon_review_data}, and Netflix competition \citep{netflix_data}. In addition to that, we test on a private Dateio dataset of transactions. There, one predicts at which merchant a bank's customer is likely to shop, in order to recommend suitable cashback offers to encourage spending.
% (We refer to all datasets by the company name; the private one being called Dateio.)

All of the public datasets contain ratings on a scale of 1-5. We normalize these values to the $[0,1]$ range. Additionally, we test on data where ratings are binarized by mapping all non-zeros to 1. Binarized Yelp data are then very similar to the Dateio data, given that items are sellers.
We prune users and items with few interactions (reviews/purchases).

To model the data distribution, working with the full input space would be inefficient. We therefore group items into the categories provided with the data, and in the case of Netflix, we group movies by their release year. In Dateio, each merchant is assigned a single category and some (more granular) tags. For example, a category might be Food \& Drink, and a tag could be American Fastfood. Values of items belonging to each group are aggregated using one of three functions: mean, sum, or disjunction. We test sum and disjunction only on the binarized variant of the data, since those functions retain the discreteness of the data, which SPNs can utilize.
% , see previous sections.

% \jiricomment{add comments on the various setups -
% mean dis sum spns,
% rating/binary,
% median optimize nospn,
% score nth,
% topk

% we only add the decrease of the score, no increase

% we only test for EASE -
% $f_j(x_i) = B_{j,:} \cdot x_i$ where $B_{j,:}$ is the $j$-th row of the weight matrix.
% we use highspy - OSS
% }
In addition to the SPN variant, we also choose how we utilize the SPN output. Following \citep{nemecek2025generating}, we test 3 variants. In the first variant, the SPN is omitted. In the other two, we either impose a constraint on the CE likelihood to be at least equal to the median likelihood of the training data, or we subtract the likelihood into the objective \eqref{eq:objective} multiplied by $\alpha = 0.1$, effectively minimizing a linear combination of $\ell_1$ distance and negative log-likelihood.

We take 50 users, and retrieve the top $k$ recommended items for each of them. Then, for each of the top three items, we generate a counterfactual input, such that the item would rank outside of the top $k$ items of a given user. We choose $k$ to be 5 or 10, which roughly corresponds to the item not being shown on a single row of recommendations, or on the first page of results.
We test the variant where the drop is guaranteed (called $k$-th) and a variant where we constrain the item's score by the original score of the $k$-th element (called score).

In terms of allowed change, we only allow the interaction values to decrease, since we strive for explanations in the form of ``you have been recommended X because you interacted with Y,'' which translates to a counterfactual where item Y is removed from the input vector. For example, recommending a movie because a person had \textit{not} seen some other movie would be non-obvious.

For the three public datasets, we do 3-fold cross-validation, training 3 recommenders for each dataset, in order for the evaluation to be less dependent on the RS performance \citep{mohammadi_are_2024}.
% For the Dateio dataset, we utilize only the production recommender as a baseline.
Throughout, we utilize EASE \citep{steck_embarrassingly_2019} in all of the data splits.
We model the formulation using the Pyomo library \citep{bynum_pyomo_2021} and solve it using an open-source solver, HiGHS \citep{huangfuParallelizingDualRevised2018}. We optimize each formulation for up to 10 minutes (a conservative limit; the median solving time of the experiments was 11 seconds). Tests on public data were run on an internal cluster with sufficient memory (up to 128GB for model training, CE generation requires notably less) and 16 CPUs, AMD EPYC 7543 or Intel Xeon Scalable Gold 6146, based on their availability.
% \jirimargincomment{runtime specs}

\subsection{Results}

% \jiricomment{mean is more difficult than sum}

We now compare the different variants of our modified LiCE for recommenders. We focus on the ability to provide globally optimal solutions and on the trade-off between plausibility and similarity.

Firstly, considering that MIO has a reputation for being slow to solve, we wish to establish how often our method obtains
% the global optimum
a solution
of the MIO formulation. In Table \ref{tab:rates}, we show the proportion of success in CE generation. Interestingly, this depends on multiple criteria. Firstly, note that without an SPN, we obtain a CE for each factual, irrespective of the input format or validity constraint ($k$-th vs score). Secondly, note that the thresholding variant fails often, except for the disjunction network. In particular, 98.8\% of the failures of the Threshold variant are cases where the model has been shown to be infeasible, likely due to the likelihood threshold being too prohibitive. %This is caused partly by the SPN format and partly by the threshold being too restrictive (nearly all failed generations are due to the formulation being infeasible), given the granularity of the trained SPN.
Thirdly, notice that using only score thresholding leads to a higher success rate, although it does not always lead to the explained item ranking beyond the $k$-th position. Indeed, approximately 30\% of CEs generated with the score threshold remain among top-$k$ recommendations.
Interestingly, this proportion is almost the same for $k$ equal to 5 and 10.
Finally, we see that the decisions of a recommender trained on ratings (rather than binarizations) seem to be more challenging to explain. %on the optimizing variant.

We can conclude that the disjunction aggregator function is the easiest to solve, and the learned SPN seems to be the least restrictive.
Thus, we restrict our focus to the disjunction aggregator on the binarized input vectors. In Table \ref{tab:stats}, we present the results on Plausibility and Similarity, measured by the log-likelihood and $\ell_1$ distance. Overall, the results correspond to the results of the original LiCE, where we see an improvement in plausibility, offset by some decrease in similarity.
% Notice especially the Dateio results with categories, where each method finds counterfactuals consisting of changing a single input, while achieving improved likelihood on average.
% \jirimargincomment{is this the own item id? It is probably caused by different number of the values over which we do the mean..}

\section{A User Study}
% We have also evaluated our method on the private Dateio dataset, using the production recommender system.
% % which would be meaningless to compare performance-wise.
% Its performance roughly corresponds to the performance on the public datasets. More importantly, we used the method to perform a study, where 20 customers were shown the explanations for up to 20 recommended merchants, constrained to a 5-position decrease in ranking.
% Each customer was presented with 5 different counterfactuals, each generated by a different variant. The customer was then asked to evaluate the subjective quality of the explanation. In the end, each subject was asked whether this improved their understanding of the model.

% CEs of all tested variants were considered understandable, and they helped strengthen users' trust in the system's reliability. The two variants considered to return the best CEs were a variant with a score decrease and no SPN, and a variant with a guaranteed 5-position drop with a disjunction SPN on samples aggregated by tags. The former was preferred because of the simplicity of the explanations, and the latter because of the ability to uncover more unexpected explanations.
% \jiricomment{describe the user-study results briefly}

% \subsection{Survey}
In addition to the numerical results above, we've performed a user study on merchant recommendation using private data from Dateio. There, the task is to suggest cash-back offers to customers, ranking the most relevant merchant offers first.
This relevance is estimated based on the user's transaction history, i.e., looking at which merchants were visited (and how often) by each customer.
The qualitative evaluation of the provided CEs is non-trivial. Transactions with suggested merchant and CE merchants may be related to various phenomena, such as the fied of business of the merchants, spatial patterns, or other user-specific shopping habits and preferences. Therefore, the best way to assess the quality of explanations is to collect feedback directly from the users, who can properly assess the relevance of counterfactual explanations. Simultaneously, we evaluated the quality of the recommender system, since it would not be efficient to discuss suggested merchants that are completely irrelevant to the user.

\begin{figure}
    \centering
    \fbox{\includegraphics[width=0.25\textwidth]{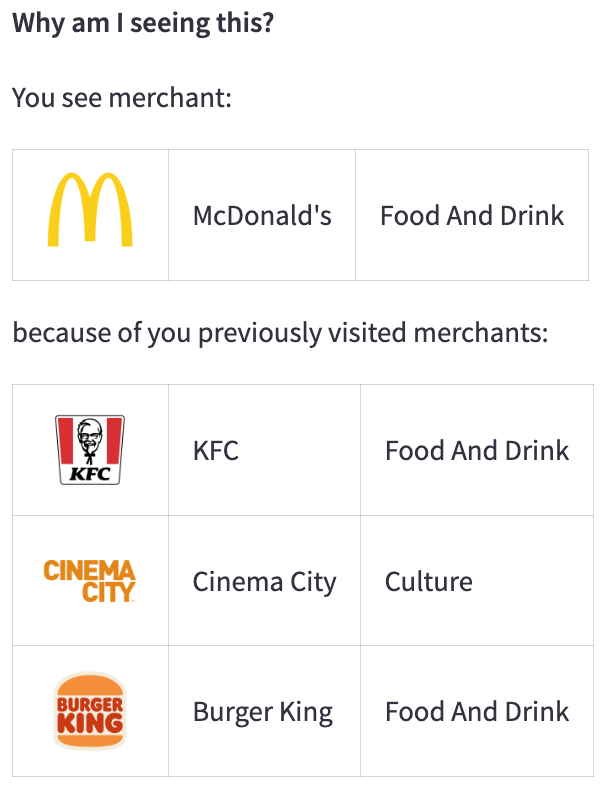}} % Adjust width as needed
    \caption{Example of a counterfactual explanation}
    \label{fig:dashboard}
\end{figure}

\begin{table}
\centering
\caption{Example of merchant suggestions using EASE.}
\begin{tabular}{clc}
\toprule
 Rank & Merchant &  EASE score \\
\midrule
    1 &         BILLA &         1.801893 \\
    2 &    McDonald's &         1.749221 \\
    3 &          Lidl &         1.700293 \\
    4 &        Albert &         1.641677 \\
\bottomrule
\end{tabular}
\label{tab:ease}
\end{table}
In order to test both the suggestions and their explanations, we created a survey that incorporates a personalized dashboard with transaction data. Merchant suggestions and CEs were generated based on the transaction history of each subject. Example ranking and counterfactual explanation can be seen in Table \ref{tab:ease} and Figure \ref{fig:dashboard}, respectively.

First, the respondent provides a general assessment of merchant suggestions, primarily considering the relevance of included merchants and their relative order. Second, the CEs of suggested merchants are evaluated. We generate multiple alternative explanations for each merchant suggestion, corresponding to different validity constraints and SPNs used. All setups are described in Table \ref{tab:ce_approaches}, where the ``score decrease'' (in C, D, and E) means setting a threshold on the score that would correspond to a drop of 5 positions in the original ranking. The 5 position drop is guaranteed in the setups A and B. Category SPN uses merchant categories (see third column in Figure \ref{fig:dashboard}) and a sum as aggregator function. Binary tag SPN is an SPN using tags (more nuanced categories, with more than one associated value allowed) and disjunction as an aggregator function. None stands for not using an SPN at all. We impose an additional restriction that the originally suggested merchant cannot appear among the counterfactual merchant(s). This forces LiCE to find a more complex explanation\footnote{Otherwise, the explanation could say that merchant $i$ was recommended since the customer has transactions with merchant $i$, which is not a desired outcome here.}.
Overall, the goal is to assess whether the LiCE can be applied in merchant recommendatoin and to choose the best configuration to find the explanations.

\begin{table}[t]
\centering
\caption{Setups used int the user study.}
\begin{tabular}{clll}
\toprule
  & Method &  Condition & SPN variant used \\
\midrule
  A & below n-th position & 5 positions drop &  binary tag SPN \\
  B &          below n-th position &        5 positions dorp & none \\
  C &    below a given score &   score decrease   &   none \\
  % C &    below a given score &   score decrease\protect\footnotemark[2]  &   none \\
  D &   below a given score &   score decrease  &  category SPN \\
  E & below a given score &  score decrease  &   binary tag SPN \\
\bottomrule
\end{tabular}
\label{tab:ce_approaches}
\end{table}
% \footnotetext[2]{Setting a score threshold that corresponds to a 5 position decrease.}

In total, 20 respondents with various demographic and socio-economic backgrounds took part in the survey, each of them evaluating approximately 10-20 merchant suggestions and corresponding counterfactual explanations. As seen in Table \ref{tab:resp_stats}, the median respondent has approximately 460 transactions within the last year, distributed among 113 merchants and 19 categories, which suggests a frequent and versatile usage of payment card.

\begin{table}[h]
\centering
\caption{Transaction history of respondents (one year period)}
\begin{tabular}{lrrrr}
\toprule
{} &  transactions &  merchants &  categories &   tags \\
\midrule
min  &      41.0 &       19.0 &         7.0 &   22.0 \\
median  &     465.0 &      113.0 &        19.0 &   92.0 \\
max  &    1903.0 &      290.0 &        22.0 &  148.0 \\
\bottomrule
\end{tabular}
\label{tab:resp_stats}
\end{table}

\subsection{Results}

Table \ref{tab:comp_stats} presents the statistics of various CE approaches. One can observe significant differences in the run-time of finding a CE. The approach C performs the fastest and provides a solution for most cases. Moreover, it produces explanations using the fewest merchants, on average. That could contribute to more concise and clear explanations. Its lack of use of an SPN would however suggest possibly lower plausibility.
\begin{table}[h]
\centering
\caption{Computation statistics}
\begin{tabular}{lrrr}
\toprule
 &  Mean duration & Proportion obtained &  Mean CE size\\
\midrule
  A &        35.261 s &          0.786 &                                     4.415 merchants \\
  B &        24.938 s &          0.774 &                                     4.037 merchants \\
  C &         \textbf{0.093 s} &          \textbf{0.905} &                                     \textbf{3.045} merchants \\
  D &        13.647 s &          0.900 &                                     3.410 merchants \\
  E &         0.312 s &          \textbf{0.905} &                                     3.503 merchants \\
\bottomrule
\end{tabular}
\label{tab:comp_stats}
\end{table}

\subsubsection{Survey results}
The evaluation of merchant suggestions\footnote{Q: Evaluate the list of recommended merchants -- How well does it correspond to your shopping preferences? (relevance of suggested merchants, their order, ..)} shows that the average respondent rates them 7.4 points out of 10. All respondents are in general satisfied and provide a positive feedback, however, some of them point out that certain merchants, especially the biggest grocery store producers, rank undesirably high.
This might be related to the fact that the current recommender does not reflect the frequency of purchases. Therefore, a global supermarket chain, which appears in the transaction history of many users and is likely to obtain a high score, might rank among the top positions even if the user only makes a single transaction with this merchant. This insight was important for further development of the recommenders.

\begin{figure}[h] % 'h' means here
    \centering
    \includegraphics[width=0.5\textwidth]{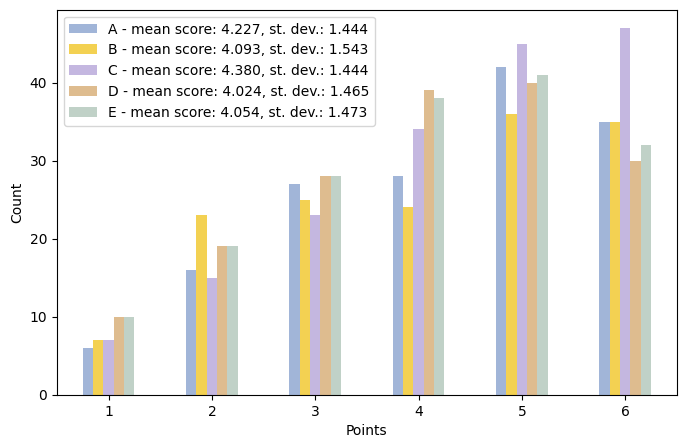} % Adjust width as needed
    \caption{Evaluation of CE explanations in terms of how sensible are they to users. Scale is from makes no sense (1) to makes prefect sense (6).}
    \label{fig:points}
\end{figure}
Considering the evaluation of CEs, the results can be found in Figure \ref{fig:points}. In general, respondents rate the explanations rather positively, as the average score\footnote{The interpretation of awarded points: 1 - 2 = does not make sense, I do not agree; 3 - 4 = ok, but I also see merchants that might not be there; 5 - 6 = makes a perfect sense.} of every approach is more than 4 points out of 6, suggesting only a few confusing explanations. Since the Friedman test shows that there exist statistically significant differences between particular approaches\footnote{$H_0$: No stat. difference between approaches; p-value = $0.0088$}, multiple Wilcoxon signed-rank tests were performed and show that approach C is significantly better than B, D, and E, based on their pairwise comparison\footnote{$H_0$: the approaches do not statistically differ; the list of adjusted p-values (applying Bonferroni correction): A vs B: 1.0, A vs C: 0.472, A vs D: 1.0, A vs E: 1.0, B vs C: 0.022, B vs D: 1.0, B vs E: 1.0, C vs D: 0.009, C vs E: 0.006, D vs E: 1.0}. From a statistical perspective, the approach A, which yields a slightly lower average score than C, can be considered equivalent.

As in the case of merchant suggestion evaluation, respondents mention a possible bias towards big merchants, which frequently appear among the CE merchants, despite having a low number of transactions. This might be resolved by redesigning the recommender system to account for this. On the other hand, respondents emphasize that when it comes to more niche merchants, the explanations work significantly better and provide more interesting information. They also appreciate the geographical patterns that the CE can provide, interconnecting seemingly unrelated merchants whose branches are, however, situated close to each other.

Based on the feedback provided, respondents prefer CEs with a lower number of merchants, since they consider them more clear and concise. However, in some cases, the CE provides a complex list of many merchants that characterize the shopping habits of the respondent. Such CE might receive a high score even though it entails multiple merchants.

\begin{figure} % 'h' means here
    \centering
    \includegraphics[width=0.4\textwidth]{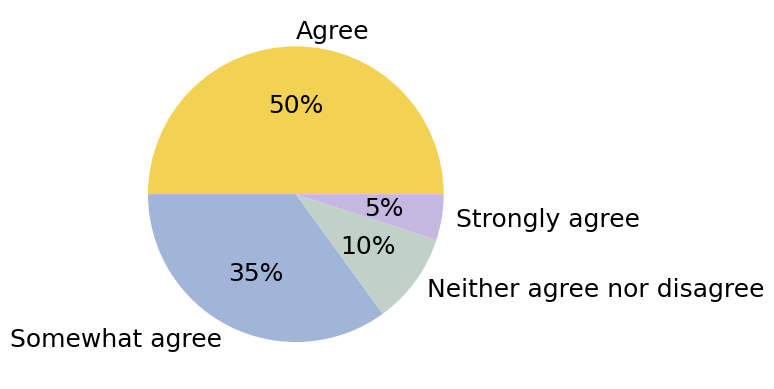} % Adjust width as needed
    \caption{Q: After seeing the explanations, I better understand the merchant-recommendation process, and I consider it more reliable.}
    \label{fig:contribution}
\end{figure}

Finally, the respondents evaluate the overall contribution of LiCE. When deciding whether they would like to see it in a banking app\footnote{Q: Would you like to have the explanation algorithm in your mobile banking app (e.g., when displaying merchants with cash-backs)?}, the responses either show unequivocal support or disapproval, resulting in the mean score of 7 points out of 10. The final question of the survey is based on the evaluation of statement presented in Figure \ref{fig:contribution}, using the Likert scale with 7 response values ranging from Strongly disagree to Strongly agree. As the figure shows, the majority of the respondents agree that the CE framework helps to make recommender decisions more transparent and understandable, whereas \textit{no one disagrees} with the statement.

% \begin{figure} % 'h' means here
%     \centering
%     \includegraphics[width=0.5\textwidth]{ce_contribution_alt.png} % Adjust width as needed
%     \caption{Contribution and usefulness of CE explanations}
%     \label{fig:contribution2}
% \end{figure}

% From the Dateio perspective, the CE results are used internally to improve the EASE recommender system and to eliminate the undesired trends reported in the feedback of the respondents. The overall results confirm that EASE is an appropriate tool for this use case and provides notable help for the further development of our product.

\section{Conclusion}
% Based on the survey, it is possible to conclude that EASE is a useful method to generate personalized merchant suggestions for users of a banking app. Moreover, LiCE is an efficient tool for explaining merchant suggestions and increasing users' trust and positive impressions. When comparing several approaches for generating CE, the simplest and most efficient method provides the best evaluations of respondents, primarily because of the simplicity of the generated explanations.

% \section{Conclusion}
We have considered the likelihood of counterfactual explanations of a recommender system and extended the LiCE method to recommender systems. We tested multiple variants and found that, especially on binary attributes, our method can reliably return explanations faithful to the underlying model. We have also supported the claims of usefulness of the method by a user study.
Assuming that the explanations were provided only upon request, even the run-time of open-source solvers may be bearable.
Finally, obtaining plausible counterfactuals could also be useful for an in-distribution data-augmentation \citep{wang_counterfactual_2021} Counterfactual Collaborative Reasoning \citep{ji_counterfactual_2023} or in studying bias in RS \citep{bhadani_biases_2021,felicioni_enhancing_2022,wang_counterfactual_2024}.

\ifCLASSOPTIONcompsoc
  % The Computer Society usually uses the plural form
  \section*{Acknowledgments}
\else
  % regular IEEE prefers the singular form
  \section*{Acknowledgment}
\fi

This work has received funding from the European Union’s Horizon Europe research and innovation programme under grant agreement No. 101070568.

% trigger a \newpage just before the given reference
% number - used to balance the columns on the last page
% adjust value as needed - may need to be readjusted if
% the document is modified later
%\IEEEtriggeratref{8}
% The "triggered" command can be changed if desired:
%\IEEEtriggercmd{\enlargethispage{-5in}}

% references section

% can use a bibliography generated by BibTeX as a .bbl file
% BibTeX documentation can be easily obtained at:
% http://mirror.ctan.org/biblio/bibtex/contrib/doc/
% The IEEEtran BibTeX style support page is at:
% http://www.michaelshell.org/tex/ieeetran/bibtex/
\bibliographystyle{IEEEtran}
% argument is your BibTeX string definitions and bibliography database(s)
\bibliography{references}

% that's all folks
\end{document}